\def\year{2022}\relax
\documentclass[letterpaper]{article} 
\usepackage{aaai22}  
\usepackage{times}  
\usepackage{helvet}  
\usepackage{courier}  
\usepackage[hyphens]{url}  
\usepackage{graphicx} 
\usepackage{natbib}  
\usepackage{caption} 
\DeclareCaptionStyle{ruled}{labelfont=normalfont,labelsep=colon,strut=off} 
\usepackage{algorithm}
\usepackage{algorithmic}

\usepackage{newfloat}
\usepackage{listings}
\floatstyle{ruled}
\newfloat{listing}{tb}{lst}{}
\floatname{listing}{Listing}
\title{Generating Real-Time Strategy Game Units Using Search-Based Procedural Content Generation and Monte Carlo Tree Search}
\author{Kynan Sorochan, Matthew Guzdial\\
}
\affiliations {
    Department of Computing Science, Alberta Machine Intelligence Institute (Amii)\\
    University of Alberta, Edmonton, Alberta, Canada\\
    ksorocha@ualberta.ca, guzdial@ualberta.ca
}

\usepackage{bibentry}

\begin{document}

\maketitle

\begin{abstract}
Real-Time Strategy (RTS) game unit generation is an unexplored area of Procedural Content Generation (PCG) research, which leaves the question of how to automatically generate interesting and balanced units unanswered. 
Creating unique and balanced units can be a difficult task when designing an RTS game, even for humans.
Having an automated method of designing units could help developers speed up the creation process as well as find new ideas.
In this work we propose a method of generating balanced and useful RTS units. 
We draw on Search-Based PCG and a fitness function based on Monte Carlo Tree Search (MCTS).
We present ten units generated by our system designed to be used in the game microRTS, as well as results demonstrating that these units are unique, useful, and balanced.
\end{abstract}

\section{Introduction}

Artificial Intelligence methods can help game developers improve their games and create new content.

In this paper we aim to show that the AI technique of Procedural Content Generation (PCG) can be implemented in Real-Time Strategy (RTS) games, particularly in the generation and balancing of new units.

Balance within RTS games can be an ongoing concern for developers.
As players invent new strategies, new imbalances within the game can be revealed.
For example, Starcraft 2 has continued to receive unit additions and balance patches for more than 10 years \cite{blizzardpatchnotes}, showing the game has still not reached perfect balance.
This indicates the process of generating interesting and balanced units is lengthy and difficult.
Using AI to help refine and speed up this process could help reach a better state in less time.

PCG unit generation has focused around NPC generation \cite{npcgeneration} and boss generation \cite{bossgeneration}
In PCG research, the majority of work done around RTS games has been automated map generation \cite{rtsmapgeneration}.
Most other RTS games research has not used PCG and has focused on creating intelligent bots and strategies \cite{reviewofrts,strategygeneration,botgeneration}.
While prior work has sought to automatically determine RTS unit balance \cite{combatbalance}, no work to our knowledge has sought to generate balanced RTS units.

In our research we attempted to create a way to utilize PCG methods to help generate new units for the RTS game, microRTS, a recognized RTS testbed developed by Santiago Ontañ{\'o}n \cite{microrts}.
We wanted these units to be unique from the pre-existing units in terms of their features and mechanics, and to keep the game balanced.
We implemented a Search-based Procedural Content Generation (SBPCG) approach with a fitness function based on an adaptation of existing work on an Monte Carlos Tree Search (MCTS)-based measure of game balance \cite{scrabblegamebalance}. 

In this paper we will show our approach for generating interesting and balanced units for microRTS.
We cover our SBPCG approach, which greedily attempts to improve an initially random unit through a measure of game balance. 
We describe the first ten generated units from our approach, as a demonstration that it can output many possible units. 
We evaluate these ten units by showing the results of having bots of varying skill levels use them, which indicates that they are both useful and balanced. 

\section{Related Work}
In this section we overview prior work done in RTS games using PCG techniques, unit and mechanics generation in video games, and strategies to determine balance within games. 
While there's been significant prior work on AI for playing RTS games \cite{botgeneration} and generating strategies to play these games \cite{strategygeneration}, we do not include a detailed discussion of these topics as they are distinct problems from the ones we focus on in this paper.

PCG has been applied previously in the domain of RTS games for map generation \cite{mapgen1,mapgen2,mapgen3}.
\cite{starcraftmapgeneration} presented work on applying evolutionary search to StarCraft map generation. Using several fitness functions, they were able to produce playable and balanced maps .
In another search-based PCG approach, \cite{generalizedmapgeneration} devised a system that takes into account objectives relating to predicted player experience. The algorithm must balance these objectives when selecting maps from the search space as some objectives are partially conflicting.
Like both of these examples, we employ search-based PCG, but with a focus on unit generation.

Unit and game mechanics generation has been researched for many years. 
\cite{chessgeneration} demonstrated a constructive-PCG approach to generate new chess-like games. Pell's program created new movement rules within specific limitations and applies them to different chess pieces. 
\cite{mikecookcog} created an evolutionary system that generates different mechanics for platform games implementing a search-based approach using code reflection. The system's fitness function chooses specific output mechanics based on their performance in game, similar to our generation process. 
\cite{conceptualexpansion} introduced the idea of conceptual expansion, a method where existing games are recombined to create new games. They tested their method by using it to recreate pre-existing games.
\cite{bossgeneration} designed a generative space of dynamic behaviors to create bosses of varying complexity for games.
Our domain differs greatly from these as we employ PCG to generate specific units and balanced mechanical effects for them, not on generating general rules or an entire game.  We additionally must take into account specific RTS-game features like currency that might not be a consideration in other types of games.

Achieving balance can be a tricky in many games. For our work we wanted to make sure that we had a way of determining that the game is still balanced once we add new units. We do not contribute new technical solutions to this problem.
In work by \cite{scrabblegamebalance}, they used simulated agents of varying levels of competency to model players playing scrabble and a simple card game. They used the win-rates of the different skill levels to approximate whether the game was balanced or not.
We draw on this approach specifically to inform our fitness function and evaluation. 
\cite{evolutionarygamedesign} employed an approach of measuring game quality through self-play simulations.
\cite{balancebehaviormodeling} attempted to model human behaviour from examples to better simulate player actions.
We identify both of these as possibilities for future work, but note that collecting enough human player data would be a large roadblock.

\section{Background}
Before we explain our system for generating RTS units, it is important to be familiar with microRTS, the platform we use in our work. microRTS, developed by Santiago Ontañ{\'o}n \cite{microrts}, was created for AI research. It is advantageous to use over other RTS games due to its simplistic design. Most other RTS-games are large and complex, having dozens of units, abilities, and other details that would greatly expand our search space. Thus a simpler domain was appropriate for this initial exploration of our approach. 

microRTS is played in an 8x8 gridspace, each player starting in opposite corners with resources, a base, a barracks, and a worker. Workers mine resources and return them to the base while players make more workers from the base and army units from the barracks. Army units can attack other units and buildings with varying power and range. Actions taken by both players occur simultaneously without pauses. The game is over once all of one of the players' units and buildings are destroyed, or the game timer reaches a pre-determined threshold. In the base game three army units can be made. A light armored melee unit, a heavy armored melee unit, and a light armored ranged unit. In our modification we generate a fourth unit with varying stats that can be made in the barracks. 

\begin{figure*}[tbh]
\centering
\includegraphics[width=6.5in]{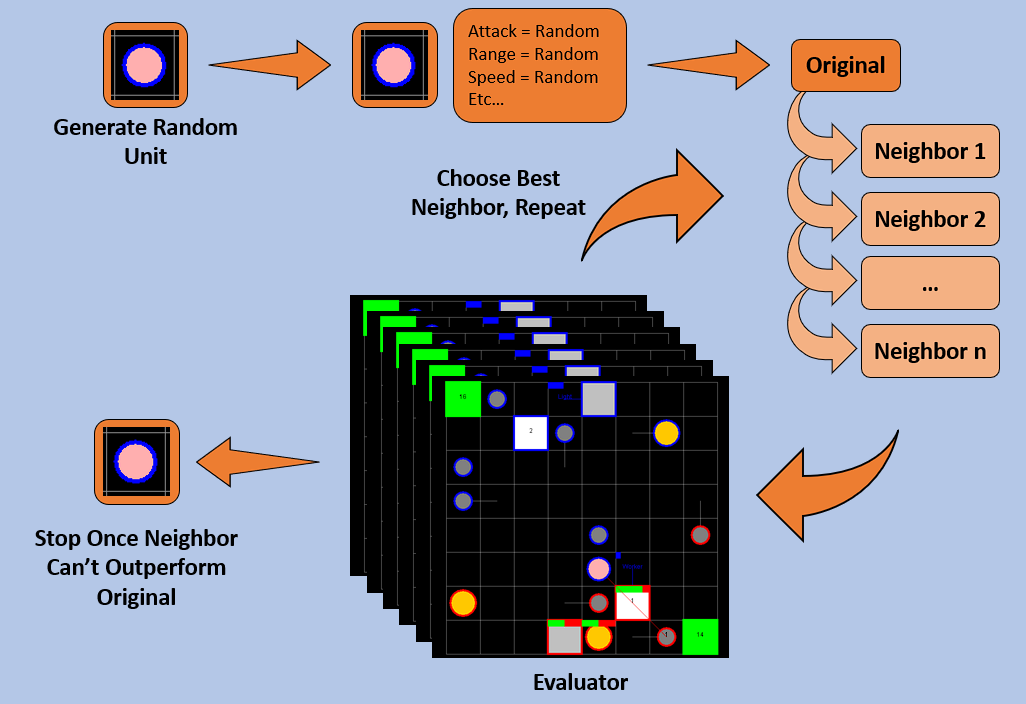}
\caption{Our search-based unit generator. First a unit is initialized at random. From there we pass it to a greedy search process, which iteratively improves the unit by selecting neighbours in the search space according to an MCTS-based fitness function.  }
\label{fig:sysoverview}
\end{figure*}

\section{System Overview}
In this section we outline our search-based generator for RTS units. 
We visualize this process in Figure \ref{fig:sysoverview}.
Our method has 3 major components: the search space that defines the unit stats and abilities, the evaluator that tests different units from the search space, and our search method that explores the search space.

\subsection{Search Space}

Our search space is defined by the different stats of units in microRTS, and our hand-authored space of mechanics defined in terms of cuases and effects.  
We did not include every unit stat in our search space, but focused on specific ones we felt were impactful for creating an interesting new unit.
The stats that made up our search space were:
\begin{itemize}
    \item Resource Cost: The number of resources needed to build the unit [1,3].
    \item HP: The number of hit points [1,4].
    \item Damage: The amount of damage done per attack [1,4].
    \item Attack Range: How many squares away the unit can attack another unit [1,3].
    \item Move Speed: How many in-game seconds are needed to move to an adjacent square [5,14].
    \item Attack Time: How many in-game seconds are needed to deal damage to an enemy unit within range [3,7].
\end{itemize}
The values above are the ranges used when generating the initial random unit. It is possible for some stats to reach outside this range while generating neighbours. It is not possible for any value to go below 1.

Outside of states, we also added a new aspect to the game: unit abilities, previously not a part of microRTS. 
Our generated units each have an ability, defined by a cause and effect.
The possible causes to trigger the ability and effects of the ability were:
\subsubsection{Causes:}
\begin{enumerate}
    \item When the unit dies
    \item When the unit takes damage
    \item When the unit deals damage
    \item When the unit attacks at least three times
\end{enumerate}

\subsubsection{Effects:}
\begin{enumerate}
    \item Return resources to the unit's player equal to cost of the player's unit for causes 1 and 2, or of the unit being attacked for causes 3 and 4.
    \item Do damage back to an attacking enemy for causes 1 and 2, or do a second attack for causes 3 and 4.
    \item Heal the unit up to 3 hit points.
    \item Double the attack speed for causes 2, 3, and 4, or cut the opponents attack speed in half for cause 1.
\end{enumerate}

\subsection{Evaluator}

Our search-based PCG approach requires an evaluator of the units to act as a fitness function. 
Our evaluator, visualized in the centre of Figure \ref{fig:sysoverview}, consists of a series of games played in which the unit can be tested.
For each call to the evaluator, a unit is evaluated over two rounds of gameplay where two Monte Carlo Tree Search bots would use the generated units.
In the first round, one bot would have access to the new unit and the other would not.
We tracked how many games the new unit was made, how much time it was alive for in each game, and how many times the bot with the new unit won the game when it made the unit.
This first round was meant to approximate the utility of the unit.
In the second round both bots would have access to the new unit.
We then tracked how many games each bot made the new unit, and in how many of those games it won.
This second round was meant to approximate the balance of the unit. 
Each round consisted of 10 games, totalling 20 games per unit.

Once the rounds were complete, the collected results would be passed through a fitness function to approximate the quality of a unit in terms of utility and balance. 
The fitness function was divided into two parts, one from each round, that would be added together to give an overall final fitness.
The higher the fitness, the better the unit overall.
In round one the purpose of the games was to determine the utility of a unit.
The most important metrics to measure this were first how many times the bot made the unit and won. If the unit was useful, it should help the bot win more times than it causes it to lose.
Second was how long the unit lasted in the game. We assume that the longer a unit exists in a game, the more of an impact it will have on the overall outcome of the game.
A higher fitness could be achieved by increasing both of these metrics.
The part of the fitness from round one was determined by
\begin{equation} \label{eq:1}
\frac{\sum_{n=1}^{\gamma}(score \pm \frac{\alpha}{\beta})}{\gamma}
\end{equation}

\noindent
where \(\alpha\) is the time the new unit was alive in game for, \(\beta\) is the total game time, and \(\gamma\) is the total number of games played where the unit was made.
The $score$ is the summed total of wins (+1) and losses (-1) of the bot in games where the unit was made.
The $\frac{\alpha}{\beta}$ is positive if the game was a win, and negative if it was a loss.
\(\gamma\) was used to normalize the round one part of the fitness.

The purpose of round two was to determine the balance of the new unit. 
In this round, both bots would have access to it. 
Therefore, we would anticipate a fifty percent winrate, unless the unit gave an unfair advantage based on player location, or one player stumbled upon an imbalanced strategy using the unit. 
If the unit allowed for an imbalanced strategy, then it is likely the player that made it first would most often win, or the player that didn't make it would never win.
The round two part of the fitness  was determined by
\begin{equation} \label{eq:2}
1-abs(0.5 - \frac{\epsilon}{\zeta + \eta})
\end{equation}

\noindent
where \(\epsilon\) is the number of games won by player 1, \(\zeta\) is the number of games player 1 made the unit, and \(\eta\) is the number of games player 2 made the unit.
This roughly equates to the win percentage of player 1. 
Initially we intended to subtract the number of games the new unit was made by both players from \(\zeta\) and \(\eta\).
However we found that the units produced still appeared to be balanced without this additional calculation, which we will show in our results later on in this paper.
If the win percentage of Player 1 is 50\%, that means it is the same for Player 2, indicating the unit keeps the game balanced.
Once the values from equations \ref{eq:1} and \ref{eq:2} are calculated they are totaled for a final fitness for each unit in the neighbor list.

\begin{table*}[tbh]
\centering
\begin{tabular}{c|c|c|c|c|c|c|c|c|c}
    \hline
    \textbf{Name} & \textbf{Cost} & \textbf{HP} & \textbf{Damage} & \textbf{Range} & \textbf{Move Time} & \textbf{Attack Time} & \textbf{Cause} & \textbf{Effect} & \textbf{Fitness} \\
    \hline
    Revenger & 3 & 4 & 2 & 3 & 13 & 10 & 1 & 2 & 1.3397\\ 
    LooTennet & 3 & 2 & 2 & 1 & 7 & 7 & 3 & 1 & 1.3773\\
    Phoenix & 2 & 1 & 3 & 1 & 15 & 3 & 1 & 1 & 1.3577\\
    Penny Pincher & 1 & 3 & 1 & 2 & 10 & 8 & 4 & 1 & 1.0068\\
    Slinger & 1 & 3 & 2 & 3 & 11 & 3 & 4 & 2 & 0.7911\\
    Statue & 1 & 4 & 1 & 3 & 11 & 5 & 4 & 1 & 0.9153\\
    Lawman & 2 & 1 & 3 & 2 & 11 & 3 & 3 & 2 & 0.8511\\
    Barrage & 1 & 3 & 2 & 4 & 13 & 3 & 4 & 2 & 0.7255\\
    Hunter & 1 & 3 & 2 & 1 & 7 & 6 & 4 & 1 & 1.0819\\
    Chopper & 2 & 1 & 2 & 2 & 7 & 5 & 2 & 4 & 0.6816\\
    \hline
\end{tabular}
\caption{Our first ten generated units, with names given by us. }
\label{tab:UnitsGen}
\end{table*}

\subsection{Search Method}
Our search method uses a greedy or hill climbing approach to find local maxima within our search space using the data collected from the evaluator step.
There could be many possible new units that could be considered interesting and balanced, so we focus on a greedy approach that can quickly find local maxima. 
Our search process began by generating a unit at random, with an arbitrary cause and effect, as discussed above. 
To generate neighbours, we'd alter the unit by incrementing and decrementing each state, and by making all possible cause and effect replacements. 
If any of the neighbour units outperformed the current unit in terms of the above fitness function, it became the new current unit and the cycle continued. 
Otherwise, we had reached a local maxima and returned the current unit.

\section{Units Generated}
We give the first 10 output units from our system in Table \ref{tab:UnitsGen}.
We did not stop our system from generating similar units, but despite this each unit is reasonably distinct. 
We assigned a name to each unit inspired by its unique traits.
Visuals of the units are not included as they all look the same within the game, an example of which can be seen in Figure \ref{fig:sysoverview}.

\begin{itemize}
    \item \textbf{Revenger}: is slow and tanky, boasting the highest HP score of 4 and the slowest attack speed of 10 in-game seconds. Its health combined with a long range of 3 makes it difficult to kill. Once killed, it exacts revenge by dealing damage to the unit that killed it, thus the name ``Revenger".
    
    \item \textbf{LooTennet}: has one of the fastest movement speeds at 7 in game seconds. Its range, HP, and damage are low, but its design isn't to kill units, but to collect their ``Loot". Every time the ``LooTennet" deals damage to an enemy unit, the owner of ``LooTennet" receives resources equal to the production cost of the enemy unit damaged.
    
    \item \textbf{Phoenix}: has only 1 HP, but it deals 3 points of damage per attack and has one of the fastest attack speeds at 3 in-game seconds. These stats allow it to kill most other units before it can be hit itself, making it a glass cannon. In the event it does get hit, it will return the resources needed to build it, allowing the player to create another ``Phoenix".
    
    \item \textbf{Penny Pincher}: is a more average unit, but it stands out in cost effectiveness. It costs only 1 resource but has a higher than average HP score of 3. It even produces resources, as its return that 1 resource every third attack.
    
    \item \textbf{Slinger}: is short for ``Gun Slinger" as this unit has a high damage output due to it being able to hit anything within a range of 3, with an attack speed of 3 in-game seconds, and deal 2 damage per hit. It also comes with the additional ability to deal double damage every third attack. This enables it to take down enemy units quickly.
    
    \item \textbf{Statue}: is similar to ``Penny Pincher" in terms of cost effectiveness and resource collection ability. It is slightly slower in movement speed, but has more HP, range, and a faster attack speed. These advantages make target selection more important, reducing the need for movement.
    
    \item \textbf{Lawman}: is most comparable to ``Slinger" with it's fast attack speed and low HP. It's cost is higher, but the trade-off is it deals more damage. Since it always attacks twice, it can instantly kill any unit in the game.
    
    \item \textbf{Barrage}: is another fast attacker with an attack speed of 3 and a long attack range of 3. It is well equipped to be an artillery-style unit with the ability to do two attacks instead of one every third attack, sending a ``Barrage" of damage toward the enemy.
    
    \item \textbf{Hunter}: is similar to the ``Penny Pincher" in ability to collect the resources of its target every third attack, but deals slightly more damage per attack. It is also faster in movement speed and attack speed, though it sacrifices some range to accomplish this.
    
    \item \textbf{Chopper}: is the only generated unit with the ability to double its attack speed when it takes damage. ``Chopper" has the fastest possible movement speed and a fast attack speed relative to the other units, enabling it to attack and move quickly, much like a helicopter.
\end{itemize}

Interestingly, none of the generated units ended up with the possible healing ability effect. This could be due to microRTS games typically being short and not allowing for unit longevity to impact the game in a significant way. Alternatively, it could be that we simply did not generate enough units, and that such a unit is in an undiscovered local optima.

\section{Evaluation}
Our goal is to produce interesting and balanced new RTS units. 
Ideally we'd have humans play with these units and judge them, but it isn't practical here.
We'd need a huge number of playtests, where each player played with each generated unit against other players with and without the unit.
This would require hundreds of games for each player, which is not possible at this point in the project.
Instead we want to automatically determine the balance and utility of our generated units.

We rely on the work of \cite{scrabblegamebalance} in which they implement an automated balance-testing method. 
They use Monte-Carlo Tree Search (MCTS) agents of varying degrees of skill.
They then have the agents play against one another, and track their win-rates.
They argue that, for a balanced game, units of the same skill level should have balanced win-rates, while agents of varying skill levels should not.
Zook et al. only used this approach in static games, we instead apply it new versions of microRTS with the new units added.
We use MCTS agents that play microRTS and then track the win percentage of these agents as well as other metrics.
We used three different skill levels, strong, medium, and weak.
The difference in strength between the agent came from their max tree depth and max iterations budget. 
Strong had the largest tree depth and iterations budget of 10 and 1000, Weak the smallest of 2 and 250, and Medium had values of 5 and 500.
We use two rounds of play for each possible agent match-up. (Ex. Strong vs Strong, Strong vs Medium, etc.)
In the first round, only player 1 has the new unit, and in the second both players have access to it.

Balance is shown by the effect the new unit has on the win-rate in the different scenarios.
When the game is balanced, evenly matched agents should have near equal win-rates, and stronger agents should have higher win-rates over weaker ones.
When a new unit is introduced and only given to one agent, it should give that agent an unfair advantage if the unit is useful.
If the unit is useful, the win percentage of the agent with it should be greater than the one it has when it has no advantage and the game is balanced.
If there is no change or even a decrease in the win percentage, then the unit is useless.
That advantage should largely disappear when both agents can build the unit, returning the win-rate to the expected value based on agent strength.

Each match-up in each round consists of 100 games per new unit. 
Across these games we tracked the number of times the new unit was made by the agent(s) that could make it, the amount of times each agent won when the unit was built, and the average amount of time the new unit survived in any game it was made.
To ensure that any change in win-rate was due to the unit, we decided that if the average amount of games the unit was made in for that round across all ten units was less than 25, we would redo the games.
This precaution was only necessary one time for the Strong vs Strong match-up, where in the first run through the average amount of games the new unit was made was 7.4.

\section{Results}
\begin{figure}[tbh]
    \centering
    \includegraphics[width=3in]{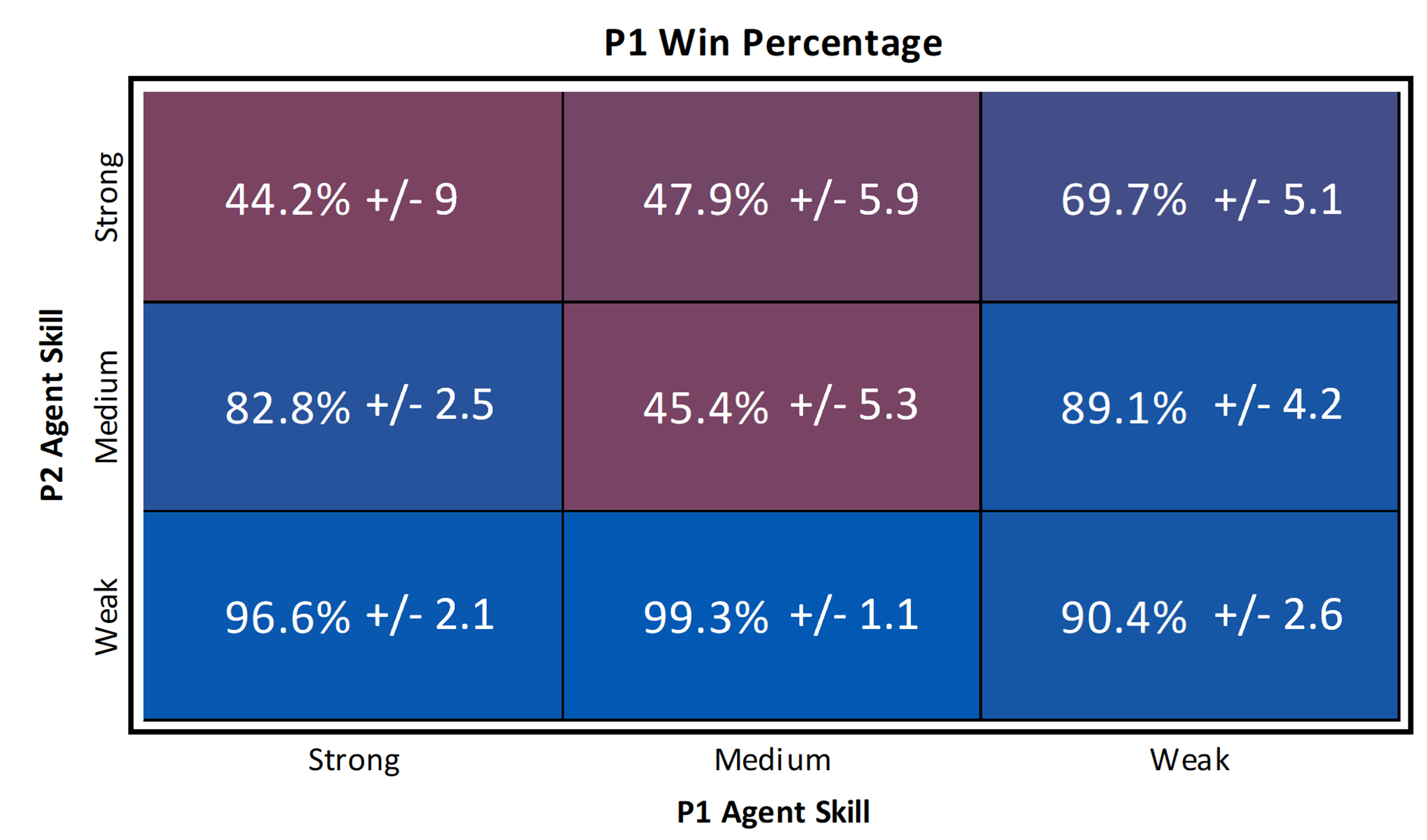}
    \caption{Average win percentage with standard deviation of Player 1 when Player 1 can build the new unit and Player 2 cannot, averaged over the 10 new units. Bluer regions correspond to a win percentage greater than 50\%. Redder regions correspond to a win percentange less than 50\%.}
    \label{fig:R1WR}
\end{figure}

We summarize the average win-rate when only one player had the new unit across all the units in Figure \ref{fig:R1WR}.
In nearly all match-ups we can see building the new unit provided an advantage. 
The only match-ups we see where the new unit did not providing an advantage is in Strong vs Strong and Medium vs Medium.
They are close to 50\%, suggesting that for these match-ups the new units had no or a slightly negative impact.
We found that this result came about due to differences in how the varying agents used the generated units.
Across the test runs the Strong agent built the new unit the least of all agents, with an average percentage of games where the unit was made across all 10 units being only 31.7\%.
For comparison Medium vs Medium made the new unit on average in 61.4\% of games and Weak vs Weak an average of 75.3\%.

The Strong agent was the default microRTS agent, with the largest search depth and iteration budget of the three agents.
All of the parameters and implementation details of this agent were chosen with the original game in mind.
Therefore, its possible that they were not a good fit to evaluate the impact of a new unit with effects well outside the bounds of the default units.
The agents with less tuned setups may have been able to avoid this in comparison.

We also found that the different agents were able to better use different units.
``Lawman" was most helpful in Strong vs Strong with a 55\% win-rate, whereas ``Hunter" had the worst win-rate at 26\%.
This doesn't necessarily mean that these are the best and worst generated units.
For example, in the Medium vs Medium match up, ``Lawman" had the worst win-rate at 39\%.
This actually supports our goal of generating interesting units by showing that our method is capable of creating units that impact the game differently at different skill levels.
This is especially important in RTS games as there is often a high skill ceiling, with different units being stronger or weaker depending on the player's skill level.

The most impressive result is the performance of the Weak agent. 
The Weak agent outperformed both the Strong and Medium agents by a significant margin.
Across the different match ups the Weak agent seemed more willing to experiment with the new units, as it usually had a higher average of amount of games where it made the unit.
This most strongly demonstrates the possible advantage given by our generated units, indicating their utility.
\begin{figure}[tbh]
    \centering
    \includegraphics[width=3in]{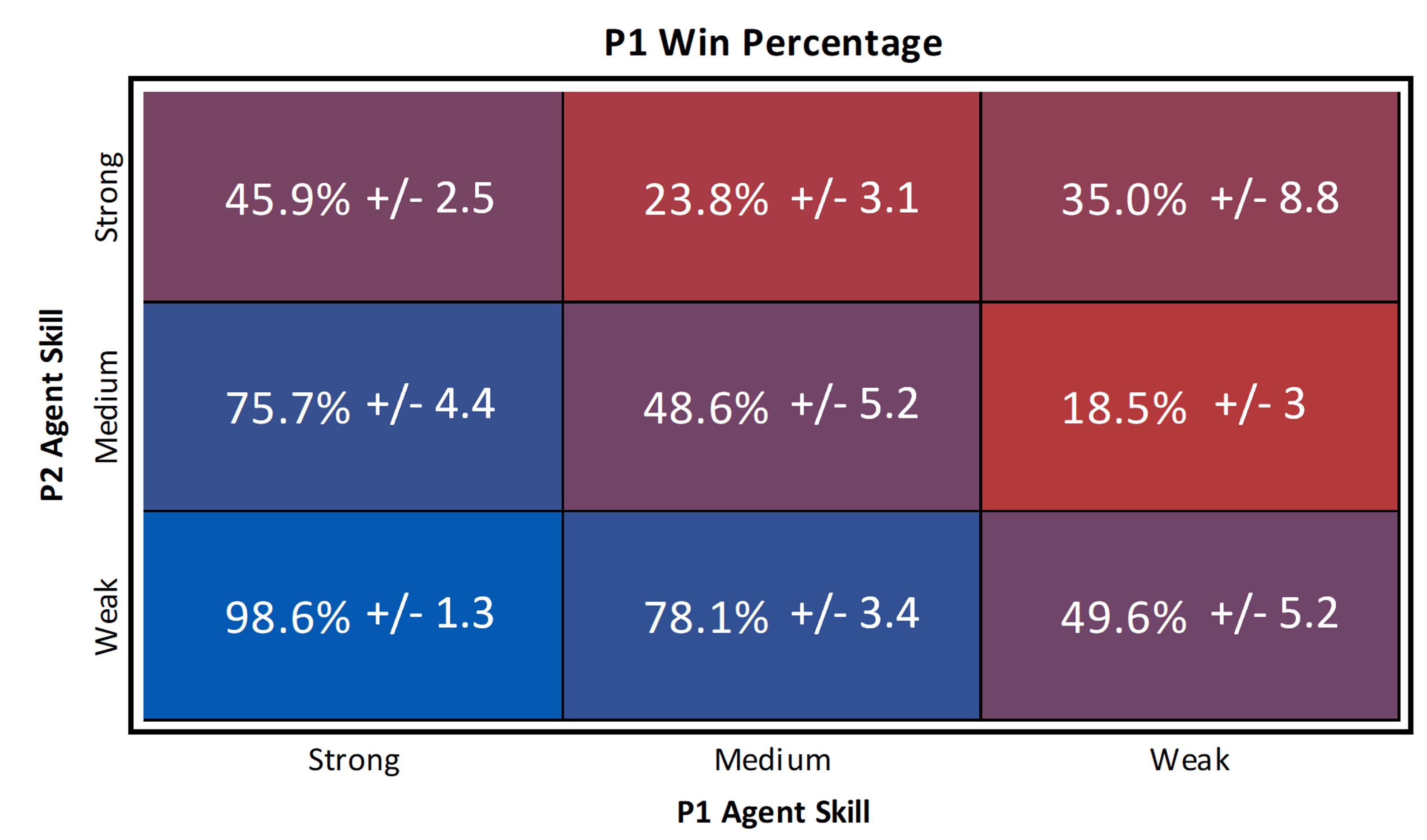}
    \caption{Average win percentage and standard deviation of Player 1 when both players can build the new unit. The x-axis indicates agent strength for Player 1, the y axis the agent strength for Player 2.}
    \label{fig:R2P1WR}
\end{figure}

Figure \ref{fig:R2P1WR} gives the average win-rate of player 1 when both agents could make the new unit.
The heat map overall displays the expected appearance for a balanced game \cite{scrabblegamebalance}, which strongly suggests that these units are overall balanced. 
Player 1's win-rate decreased compared to Figure \ref{fig:R1WR} and vice versa for match-ups where player 1 is weaker, while the even match-ups moved closer to 50\%.
These changes from Figure \ref{fig:R1WR} show the advantaged that was granted by only one player having the new unit.
The only match-up where we do not see the expected effect is in the case where Player 1 is Strong and Player 2 is Weak.
While Strong still beats Weak as expected, Weak is more frequently winning than in the Strong vs Medium match-up. 
We anticipate this is due to Weak's willingness to make the new units, compared to Strong's unwillingness. 

Another interesting result is the drastic swing in win percentages in some match-ups, for example in Weak vs Medium.
This and the Weak vs Strong result shows that some of the new units might be a little ``cheap" in terms of giving lower skill players more of an advantage.
However, given that overall the trend follows our expectations, we anticipate that on average the units are balanced. 

\section{Limitations and Future Work}
Our goal in this work was to create a method of generating new, interesting, and balanced units in an RTS game.
From our results we conclude that our search-based PCG method was successful.
We created unique new units that generally gave a player an advantage when only one player could make them and that kept the game balanced when both players could make them.

We believe that this method could be expanded and refined by addressing some limitations we encountered during our research.
For example, in our experiment we had the agents play on the standard map that is used in the microRTS competition.
While it doesn't seem that any of our units were map-specific, using a variety of maps of different sizes and qualities, could lead to generating different units. 

It can be difficult to create automated agents that play RTS games like a human would.
As such, it's unclear if our units would be balanced if used by humans. 
We didn't use human playtesting, but would like to in the future.
Having human feedback would prove much more valuable in creating units that make the game more interesting for humans.
Alternatively, the automated agents could be further improved to play more like humans.
With examples of prior human playtraces we could draw on approaches like \cite{balancebehaviormodeling} to attempted to bias our agents to behaving more like humans. 

During training, we only ran each round for ten games. 
This was largely due to limited computation power, which restricted the number of games played and the parameters of our MCTS agents. 
More games with more varied agents could potentially increase the balance of the units.

In the future, we hope to expand this work and apply it to a larger scale.
Applying and improving this method to work on a more complex game such as StarCraft Broodwar or StarCraft 2 would be required to prove if this is an approach that could work in an industry setting. 
However, a larger game with more features would lead to high computation power costs.
The way of testing units and calculating balance would likely need to be improved.
Something such as a Deep Reinforcement Learning agent could help solve some of these problems, but it would also introduce new ones. 
For example, training would be a problem since each new unit would functionally alter the MDP.
Another way of incorporating our work into a larger game would be to use it in areas other than a 1v1 competitive game mode.
Cooperative play or campaign modes where humans play against the computer are areas where new and unique content is valuable and balance is less of a concern.
Using our technique with more of a focus on generating unique units could allow for new types of RTS game designs.

\section{Conclusions}
In this paper, we have shown how it is possible to use Search-Based PCG and MCTS to generate new units for RTS games.
For the first time, we defined the problem of how to generate balanced and useful RTS units.
Our method relies on a fitness function designed to evaluate the quality of a generated unit based off the performance results of MCTS agents playing games against one another with and without the unit.
We presented ten unique and balanced units generated for microRTS and compared them against each other.
We evaluated these units in a large-scale MCTS study and found evidence for their utility and balance. 
This provides evidence to the utility of our problem definition and approach for the task of RTS unit generation.

\section{Acknowledgments}
We acknowledge the support of the Alberta Machine Intelligence Institute (Amii).
\bibliography{aaai22}

\end{document}